\documentclass[conference]{IEEEtran}
\IEEEoverridecommandlockouts
\usepackage{titlesec}
\usepackage{graphicx}
\usepackage{cite}
\usepackage{circledsteps}
\usepackage{amsmath}
\usepackage{amssymb}
\usepackage{amsfonts}
\usepackage{amsthm}
\usepackage{hhline}
\usepackage{booktabs}
\usepackage{stfloats}
\usepackage{bm}
\usepackage{verbatim}
\usepackage[utf8]{inputenc}
\usepackage{subfigure}
\usepackage{array}
\usepackage{color}
\usepackage{algorithm}
\usepackage{algpseudocode}
\usepackage{framed}
\usepackage{subeqnarray}
\usepackage{cases}
\usepackage{epstopdf}
\usepackage{float}
\usepackage{upgreek}
\usepackage{lettrine}
\usepackage{multirow}
\usepackage{multicol}
\usepackage{arydshln}
\usepackage{mathrsfs}
\usepackage{xcolor}
\usepackage{mdframed}
\usepackage{comment}
\usepackage{threeparttable}
\usepackage{footnote}
\usepackage[hidelinks]{hyperref}
\usepackage{setspace}
\usepackage{bbding}

\graphicspath{{images/}}

\def\BibTeX{{\rm B\kern-.05em{\sc i\kern-.025em b}\kern-.08em
    T\kern-.1667em\lower.7ex\hbox{E}\kern-.125emX}}
    
\begin{document}
	\title{QoS-Aware Token Scheduling and Private Data Valuation for Multi-Modal Agentic Networks}
  		\author{\IEEEauthorblockN{Yao~Du\IEEEauthorrefmark{1}\IEEEauthorrefmark{2}, Jing~Liu\IEEEauthorrefmark{1}, Pengfei~Xu\IEEEauthorrefmark{2},  Zehua~Wang\IEEEauthorrefmark{1}\IEEEauthorrefmark{3}, Victor~C.M.~Leung\IEEEauthorrefmark{1}\IEEEauthorrefmark{4}\IEEEauthorrefmark{6}, Cyril~Leung\IEEEauthorrefmark{1}, Victoria~Lemieux\IEEEauthorrefmark{1}}
  		\IEEEauthorblockA{\IEEEauthorrefmark{1}The University of British Columbia, \IEEEauthorrefmark{2}LazAI Network}
  		\IEEEauthorblockA{\IEEEauthorrefmark{3}China University of Mining and Technology, \IEEEauthorrefmark{4}SMBU, \IEEEauthorrefmark{6}Shenzhen University}
  		\thanks{\scriptsize This is the accepted manuscript of a paper to appear in the 2026 IEEE International Conference on Multimedia \& Expo (ICME), Bangkok, Thailand, July 5–9, 2026.  The final version will be available in IEEE Xplore. \copyright~2026 IEEE. Personal use of this material is permitted. Permission from IEEE must be obtained for all other uses.}
  		\thanks{\scriptsize This work was supported in part by Mitacs Project IT47821 under Grant QJLI GR037230, the Natural Sciences and Engineering Research Council (NSERC) of Canada under Grants RGPIN-2019-06348, RGPIN-2020-05410, RGPIN-2021-02970, DGECR-2021-00187, the Guangdong Pearl River Talents Recruitment Program Grant 2019ZT08X603, the Guangdong Pearl Rivers Talent Plan Grant 2019JC01X235, the Mitacs Project IT44479, and the UBC PMC-Sierra Professorship in Networking and Communications (Corresponding authors: Zehua Wang and Jing Liu).}
  		}
	\maketitle
	
	\begin{abstract}
        In agentic systems, human-generated data records anchor the value of AI services. Yet cloud compute pipelines centralize processing on remote servers. Data centralization reduces personal data sovereignty and may potentially degrade the quality of service (QoS). Meanwhile, user contributions are diverse in quantity and quality: decentralized records can be \mbox{biased}, noisy, and heterogeneously distributed. To address the data challenge, we study fair token allocation and private data valuation for decentralized and resource-constrained agentic systems. Our approach embeds multi-modal representations in a shared semantic space and releases differentially private (DP) prototypes to preserve utility while reducing semantic leakage. With the DP guarantee, we design a fair token allocation scheme that rewards effective contributions and remains robust to data heterogeneity and AI resource scarcity. Extensive simulations demonstrate improved contribution-based fairness and QoS compared to standard benchmarks. The improved resistance to image reconstruction attacks indicates enhanced privacy for multi-modal personal data.
	\end{abstract}

	\section{Introduction}\label{sec: introduction}   
    Nowadays, artificial intelligence (AI) is increasingly enabling applications at the network edge~\cite{10599391}. Agentic applications, powered by multiple AI agents, have become a major research focus. Traditionally, these systems follow a cloud-centric pipeline: user data are collected on-device, transmitted to remote servers, processed by large foundation models in the core network, and then returned as agent actions. While this architecture accelerated early AI adoption, it also introduces substantial privacy and security risks because sensitive data must leave the device. For instance, voice assistants may upload continuous audio snippets, and recommendation agents may infer or expose private attributes (e.g., financial records) without explicit user consent.

    As AI decisions become more autonomous and financially consequential, there is a growing shift toward edge intelligence, where AI agents execute closer to users on personal devices or edge servers. This shift reduces latency and reliance on centralized infrastructure, and it better supports user data sovereignty. However, decentralized edge intelligence~\cite{du2024towards} introduces new systems challenges: user contributions are heterogeneous, resources such as compute and AI quotas are limited, and an ad hoc scheduler can waste or amplify unfair allocation of scarce edge resources. The trade-off between quality of service (QoS) and data sovereignty leads to a central question:
    
    \textbf{How can we quantify data value in a privacy-preserving way and fairly allocate scarce AI usage quotas?}

    
    While the Shapley value is a well-studied tool for data valuation~\cite{rozemberczki2022shapley}, Shapley-based schemes are often impractical in decentralized settings. Computing data Shapley values typically requires repeated model retraining and external access to raw data. Moreover, privacy-preserving cross-validation~\cite{du2024towards} can provide reliable estimates, but it assumes the availability of a local validation set, which adds the deployment complexity of agentic applications. Furthermore, trusted execution environments~\cite{kato2023olive} offer efficient integrity and confidentiality for outsourced computation, but depend on hardware trust and can be vulnerable to side channels. In contrast, zero-knowledge proofs~\cite{chen2024zkml} can prove correct computation validation without revealing inputs or AI models, but often incur high proof-generation overhead on resource-constrained edge devices.
    
    To address this gap, we study differential privacy (DP) for private data valuation and token-bucket scheduling for fair AI quota allocation. In decentralized multi-modal agentic systems, we embed user contributions into a shared semantic representation space and cluster embeddings into individual-centric decentralized autonomous organizations (iDAOs)~\footnote{https://lazai.network/learn/idao} for decentralized data valuation. Building on standard DP guarantees~\cite{dwork2014algorithmic}, we release DP-protected prototypes as iDAO catalog entries to support utility while limiting semantic leakage under reconstruction attacks. More concretely, our contributions are:
    \begin{itemize}
        \item We introduce a QoS-aware incentive framework that treats scarce AI compute resources, rather than static payments, as the dynamic currency for decentralized markets. By implementing data anchoring tokens (DAT)~\footnote{https://github.com/0xLazAI/contracts}, we settle contribution value into the parameters of a token-bucket scheduler, ensuring resource access is mathematically proportional to contributed data utility.
        \item We propose a semantic market primitive that solves the ``discovery-privacy" paradox. By representing raw data as iDAO-governed, DP-protected prototypes, we enable agents to semantically search and trade knowledge with a formal DP guarantee, thereby eliminating the need to expose raw data during the valuation phase.
        \item We provide extensive evaluations demonstrating that the token allocation scheme maintains fair allocation even under highly unbalanced contribution distributions. To validate the system's utility in adversarial environments, we empirically prove the system’s superior resistance to reconstruction attacks compared to baselines.
    \end{itemize}

    \section{System Model and Problem Formulation}\label{sec: system_model} 
    In this section, we introduce the system model of data valuation in the context of verifiable agentic systems. We further map the data valuation problem to a token-bucket scheduling problem to improve the fairness and QoS in decentralized agentic systems.

    \subsection{Agentic Network Model}
    \begin{figure}[ht]
		\centering
		\includegraphics[width=0.5\linewidth]{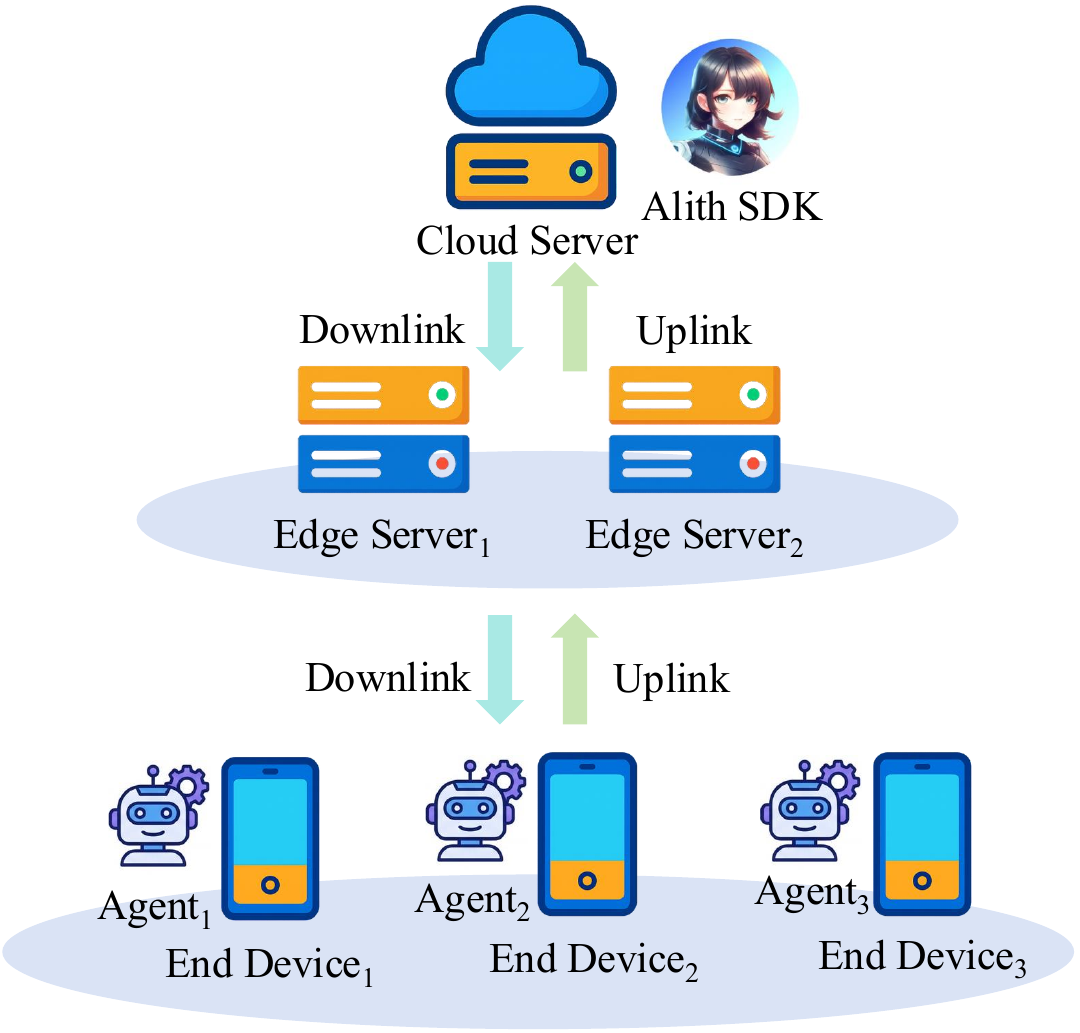}
		\caption{Cloud-edge-end collaborative agentic framework. An example with $M=2$ and $N=3$.}
		\label{fig: end_edge_cloud}
	\end{figure}
    
    As shown in Fig.~\ref{fig: end_edge_cloud}, we consider an end–edge–cloud collaboration architecture for the agentic network. Specifically, AI agents $n\in\mathcal{N}=\{1,2,\ldots, N\}$ run at the network edge and interact with end users. When local resources are insufficient, an agent offloads part of its workload to distributed edge servers $m\in\mathcal{M}=\{1,2,\ldots, M\}$ that collectively form a decentralized AI compute network. Beyond providing shared compute, edge servers conduct Layer-2 coordination, i.e., off-chain validation of submitted data and computation outcomes, generate verifiable proofs, and periodically commit summaries to the on-chain ledger for reward settlement (e.g., via DAT). We also consider a remote cloud server operated by a large AI provider as a fallback for tasks that cannot be efficiently executed at the edge.
    
    \subsection{Data Valuation Model}
    We adopt the DAT as an on-chain abstraction for valuing data and proofs in our agentic system. A DAT is a semi-fungible token designed for AI-native digital assets (datasets, models, or computation results); each token jointly encodes an ownership certificate, usage right (quota), and value share for future revenue.
    
    For any agent $n \in \mathcal{N}$, $u_{n} \in \mathbb{R}_{\ge 0}$ denotes the allocated utility quota. Let $V$ denote the total utility quota available for allocation. Then,
    \begin{equation}
    \sum_{n=1}^{N}u_{n} \leq V. \label{equ:val_sum}
    \end{equation}

    For any agent $n\in\mathcal{N}$, let $v_{n}\geq0$ represent the DAT \emph{data value} governing revenue splits. We define the DAT record as $\big({\text{address}}_n, v_{n},\, \rho_{n}\big)$, where $\text{address}_n$ denotes the ownership proof linked to an account address and $\rho_{n}$ is a compact metadata pointer (e.g., integrity hash and provenance proof). Let $\mathbf{v} = (v_{1}, v_{2}, \dots, v_{n})$ and $v_{n}\leq f(\mathbf{d}_n)$, where $\mathbf{d}_n$ denotes the data contribution by agent $n$ and $f(\mathbf{d}_n)$ is the valuation function calculating the data value of $\mathbf{d}_n$. Details of the non-negative function $f(\mathbf{d}_n)$ are introduced in Section~\ref{sec: novelty_valuation}. Therefore, we have
    \begin{equation}\label{cons:share}
        0 \leq v_{n} \leq f(\mathbf{d}_n).
    \end{equation}
    
    In Section~\ref{subsec: token_bucket_schedule}, $v_{n}$ is used to parameterize the refill rate of the token-bucket scheduler. By using a personalized refill rate, higher-valued contributors receive proportionally larger and more frequent AI quotas.

    \subsection{Threat Model}
    We assume a partially trusted agentic network: the ledger and smart contracts execute correctly, but individual DAT contributors may behave strategically.

    We consider that servers, including data validators and edge computing nodes, may be curious about personal information and could mount membership inference attacks. In Section~\ref{sec: token_allocation}, differential privacy noise is applied for privacy protection. Let $\mathcal{M}$ denote a randomized mechanism that takes a dataset as input and outputs a result. Let $D$ and $D'$ denote two neighbouring datasets that differ in at most one individual's record. Let $S$ denote an arbitrary measurable subset of the output space of $\mathcal{M}$. Let $\varepsilon \ge 0$ denote the privacy budget (smaller $\varepsilon$ means stronger privacy), and let $0 \le \delta < 1$ denote the probability of a privacy violation.
    
    A randomized mechanism $\mathcal{M}$ is $(\varepsilon,\delta)$-differentially \mbox{private~\cite{dwork2014algorithmic}} if for all neighbouring datasets $\mathbf{d}, \mathbf{d}'$ and all measurable sets $S$,
    \begin{equation}
    \Pr[\mathcal{M}(\mathbf{d}) \in S]
    \le e^{\varepsilon} \Pr[\mathcal{M}(\mathbf{d}') \in S] + \delta,
    \end{equation}
    and symmetrically,
    \begin{equation}
    \Pr[\mathcal{M}(\mathbf{d}') \in S]
    \le e^{\varepsilon} \Pr[\mathcal{M}(\mathbf{d}) \in S] + \delta.
    \end{equation}
    
    \subsection{Fairness Metrics}\label{model:fairness}
    To quantify the fairness of quota or resource allocation among $N$ agents, we employ two classical inequality measures: Jain's fairness index and the Gini coefficient.
    
    Let $\mathbf{u} = (u_{1}, u_{2}, \dots, u_{n})$ denote the non-negative utility quota (i.e., effective AI quota) received by each agent. Let $\mathbf{r} = (r_{1}, r_{2}, \dots, r_{n})$ denote the reward rate, 
    \begin{align}
        r_{n} =
        \begin{cases}
            \frac{{u}_n}{{v}_n}, &v_n > 0,\\
            0, &v_n = 0.
        \end{cases}
    \end{align}

    Our fairness analysis focuses on the distribution of reward rates, i.e., reward per contribution. \textbf{Agents with similar data contributions should have comparable reward rates}.
    
    Additionally, ${J}_{\text{min}}\in(0,1)$ and ${G}_{\text{max}}\in(0,1)$ represent the fairness bounds to mitigate algorithmic bias among agents and centralized dominance over system utility allocation.
    
    \subsubsection{Jain's Fairness Index.}
    Jain's index $J(\mathbf{r})$ evaluates how evenly the reward rates are distributed across agents.
    \begin{equation}
    J(\mathbf{r}) = \frac{\left( \sum_{n=1}^{N} r_{n} \right)^2}{N \sum_{n=1}^{N} r_{n}^2},
    \end{equation}
    where
    \begin{equation}\label{cons:Jain}
    \quad J(\mathbf{r}) \geq {J}_{\text{min}}, 0 < {J}_{\text{min}} \leq 1.
    \end{equation}
    Note $J(\mathbf{r}) = 1$ if and only if all agents receive the same reward rate. A smaller $J(\mathbf{r})$ reflects increasing disparity in the token allocation.

    \subsubsection{Gini Coefficient.}
    In contrast, the Gini coefficient $G(\mathbf{r})$ captures the pairwise deviations.
    \begin{equation}
    G(\mathbf{r})
    = \frac{\displaystyle \sum_{i=1}^{N} \sum_{j=1}^{N} 
        \left| r_{i} - r_{j} \right|}
         {2 N \displaystyle \sum_{n=1}^{N} r_{n}},
    \end{equation}
    where
    \begin{equation}
    \quad G(\mathbf{r}) \leq {G}_{\text{max}}, 0 \leq {G}_{\text{max}} < 1.\label{cons:Gini}
    \end{equation}
    Note that $G(\mathbf{r}) = 0$ corresponds to a perfectly equal reward rate distribution and larger values indicate higher inequality.

    \subsection{Joint Incentivization Problem}
    We acknowledge that the fairness metrics are rarely considered in speculative token markets. In contrast, decentralized AI computing infrastructures, such as the LazAI network\footnote{https://lazai.network/}, must effectively allocate AI quotas to prevent service outages caused by resource monopolization.

    In the context of decentralized AI computing networks, we model the social welfare function, denoted by $\mathcal{F}$, as follows:
    \begin{equation}\label{equ:social_welfare}
        \mathcal{F}(\mathbf{v}, \mathbf{u}) = \sum_{n=1}^{N} v_{n} u_n \leq \sum_{n=1}^{N} f(\mathbf{d}_n) u_n.
    \end{equation}
    One observation from the above objective is that $\mathcal{F}$ encourages valuable data contribution to maximize $f(\mathbf{d}_n)$. For $n\in\mathcal{N}$,
    \begin{align}\label{prob:token_allocation}
    \mathop{\text{maximize}}_{\mathbf{v}, \mathbf{u}}&\quad \mathcal{F}(\mathbf{v}, \mathbf{u}), \\
    \text{s.t. }
    &\quad (\ref{equ:val_sum}), (\ref{cons:share}), (\ref{cons:Jain}), (\ref{cons:Gini}). \nonumber
    \end{align}

    We note that (\ref{prob:token_allocation}) is non-convex due to the objective function (\ref{equ:social_welfare}) and constraints (\ref{cons:Jain}), (\ref{cons:Gini}). Therefore, it is very challenging to solve (\ref{prob:token_allocation}) optimally. In the following sections, we adopt a two-stage approach: we first optimize $\mathbf{u}$ given $\mathbf{v}$ in Section~\ref{sec: token_allocation}; then, we optimize $\mathbf{v}$ given $\mathbf{u}$ in Section~\ref{sec: novelty_valuation} to obtain an effective solution for decentralized AI compute networks.
    
    \section{QoS-Aware Token Allocation}\label{sec: token_allocation}
    In this section, we explore a token allocation scheme that balances fairness and performance. Given $\mathbf{v}$, we show that the best strategy for each agent under this scheme is to actively use the remaining quota. We embed a ``spend to earn more" property in our incentive design. That is, no additional tokens will be allocated to an agent with a full token bucket.

    \subsection{Problem Reformulation}
    Given $\mathbf{v}$, (\ref{prob:token_allocation}) is reformulated as
    \begin{align}\label{prob:token_bucket_allocation}
    \mathop{\text{maximize}}_{\mathbf{u}}&\quad \sum_{n=1}^{N} v_{n} u_n,\\
    \text{s.t. }
    &\quad (\ref{equ:val_sum}), (\ref{cons:Jain}), (\ref{cons:Gini}). \nonumber
    \end{align}
    Note that (\ref{prob:token_bucket_allocation}) is still non-convex due to constraints (\ref{cons:Jain}) and (\ref{cons:Gini}). In low-latency computing scenarios, non-convex optimization problems are typically handled by seeking near-optimal solutions that significantly reduce optimization complexity. To enable real-time token allocation in agentic systems, we therefore propose a simple yet effective scheme based on token-bucket scheduling.

    \subsection{Token-Bucket for Agents}\label{subsec: token_bucket_explain}
    Token-bucket scheduling regulates network traffic by accumulating tokens in a limited-capacity bucket to control data bursts while enforcing a reliable long-term service~\cite{shan2021towards}. In a decentralized AI system deployed at the network edge, computing resources are similarly constrained compared with bandwidth-limited networks. Therefore, it is critical to design a mechanism that uses AI quota efficiently while avoiding service outages due to resource limitations.

    \begin{figure}[ht]
		\centering
		\includegraphics[width=0.4\linewidth]{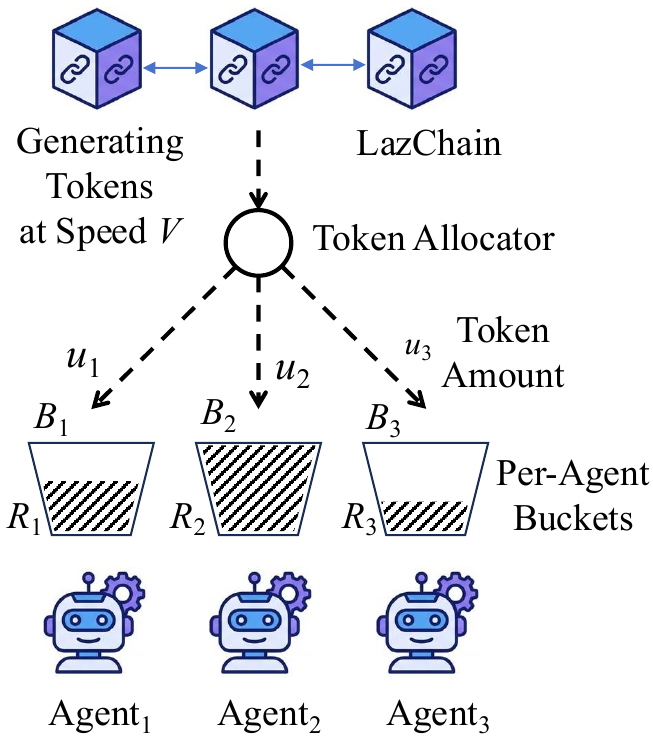}
		\caption{Token-bucket scheduling in a three-agent system. Utility quotas are represented as tokens capped by per-agent bucket sizes. Blockchain servers read the DAT information to distribute tokens.}
		\label{fig: token_buckets}
	\end{figure}

    As illustrated in Fig.~\ref{fig: token_buckets}, we adopt the token buckets for agentic systems by treating \emph{AI quota} as tokens in a virtual bucket for each agent. By using a virtual ``bucket" to store tokens as utility quotas, agents are allowed to have bursts of inference queries up to the bucket's capacity. Let $B_n$ and $R_n$ respectively denote the bucket size and remaining tokens for agent $n\in\mathcal{N}$. Typically, $B_n$ corresponds to service subscription tiers. Then, we have
    \begin{align}\label{cons:bucket_constraint}
        0 \leq u_n \leq B_n - R_n.
    \end{align}

    \subsection{QoS-Aware Token Scheduling}\label{subsec: token_bucket_schedule}
    As described above, token-bucket scheduling is a conventional network traffic shaping method~\cite{shan2021towards}, which we utilize to reshape per-agent AI quota allocation. Based on the token-bucket model, we relax (\ref{prob:token_bucket_allocation}) to a token allocation problem as follows:
    \begin{align}
    \mathop{\text{maximize}}_{\mathbf{u}}&\quad \sum_{n=1}^{N} v_{n} u_n,\\
    \text{s.t. }
    &\quad (\ref{equ:val_sum}), (\ref{cons:bucket_constraint}). \nonumber
    \end{align}

    For any agent $n\in\mathcal{N}$, we can obtain the optimized token allocation as follows: 
    \begin{align}\label{equ:user_token}
        u_{n} =
        \begin{cases}
            \min\bigl\{\frac{v_{n}}{\sum_k v_k} V,\; B_n - R_{n} \bigr\} , &R_n < B_n,\\
            0, &R_n = B_n.
        \end{cases}
    \end{align}

    Therefore, the optimal strategy for a rational agent is to actively spend its tokens on useful inference and to keep contributing high-quality data (Section~\ref{sec: novelty_valuation}), rather than hoarding tokens or remaining idle. We further provide simulation results and analysis in Section~\ref{sec: data_results} to justify the effectiveness of our proposed schemes.
    \section{Differential Private Data Valuation}\label{sec: novelty_valuation}
    Once we obtain the token allocation strategy $\mathbf{u}$, the next stage is to derive an effective data valuation function $f(\mathbf{d}_n)$ to calculate $\mathbf{v}$ fairly and authentically. In decentralized AI systems, a key research and engineering question is: how can data value be evaluated accurately without compromising privacy? In this section, we explore a multi-modal agentic novelty detection method with differential privacy guarantees to answer the above question.

    \subsection{Problem Reformulation and Analysis}
    Given $\mathbf{u}$, we relax and reformulate (\ref{prob:token_allocation}) as
    \begin{align}\label{prob:data_valuation}
    \mathop{\text{maximize}}_{\mathbf{v}}&\quad \sum_{n=1}^{N} v_{n} u_n,\\
    \text{s.t. }
    &\quad (\ref{cons:share}). \nonumber
    \end{align}
    Note (\ref{prob:data_valuation}) can be decomposed into and equivalent to $N$ independent linear programming problems.
    
    For any $n\in\mathcal{N}$,
    \begin{align}\label{prob:decomposed_data_valuation}
    \mathop{\text{maximize}}_{v_{n}}&\quad v_{n} u_n,\\
    \text{s.t. }
    &\quad (\ref{cons:share}). \nonumber
    \end{align}
    Then, we can obtain the solution for agent $n$ as follows:
    \begin{align}\label{equ:share_raito}
        v_{n} =
        \begin{cases}
            f(\mathbf{d}_n), &u_n > 0,\\
            0, &u_n = 0.
        \end{cases}
    \end{align}
    To obtain the best data value $v_{n}$, each agent must contribute $\mathbf{d}_n$ to maximize $f(\mathbf{d}_n)$. Therefore,
    \begin{equation}\label{equ:optimal_data}
        \mathbf{d}^{\star}_n = \arg\max_{\mathbf{d}_n \in \mathcal{D}_n} f(\mathbf{d}_n),
    \end{equation}
    where $\mathcal{D}_n$ denote all possible data contributions from agent $n$. Equation \eqref{equ:optimal_data} implies that, under the proposed system, a rational agent will contribute as much as possible to maximize the data value computed by $f(\mathbf{d}_n)$.
    \subsection{Private Data Valuation}
    Let $| \cdot |$ denote the set cardinality. Then, the data quantity of $\mathbf{d}_n$ can be represented as $|\mathbf{d}_n|$. We further define a novelty score function, denoted by $\phi(\mathbf{d}_n)\in [0,1]$, to quantify the average data novelty over $\mathbf{d}_n$. We propose
    \begin{align}\label{equ:data_valuation}
        f(\mathbf{d}_n) = \phi(\mathbf{d}_n)\ln(1+|\mathbf{d}_n|).
    \end{align}

    We consider a dynamic data valuation function that depends on both data quality and data quantity. The main indicator of data quality, $\phi(\mathbf{d}_n)$, measures how the contributed data $\mathbf{d}_n$ enriches the discovered knowledge of the decentralized agentic system. By definition, noisy contributions yield $\phi(\mathbf{d}_n) = 0$. In contrast, the natural logarithm is used to capture the diminishing marginal utility of increasing data quantity~\cite{du2024towards}. In this paper, we propose a data novelty-based valuation method to approximate the true data value with reduced computational complexity.
    
    \begin{figure}[ht]
		\centering
		\includegraphics[width=0.5\linewidth]{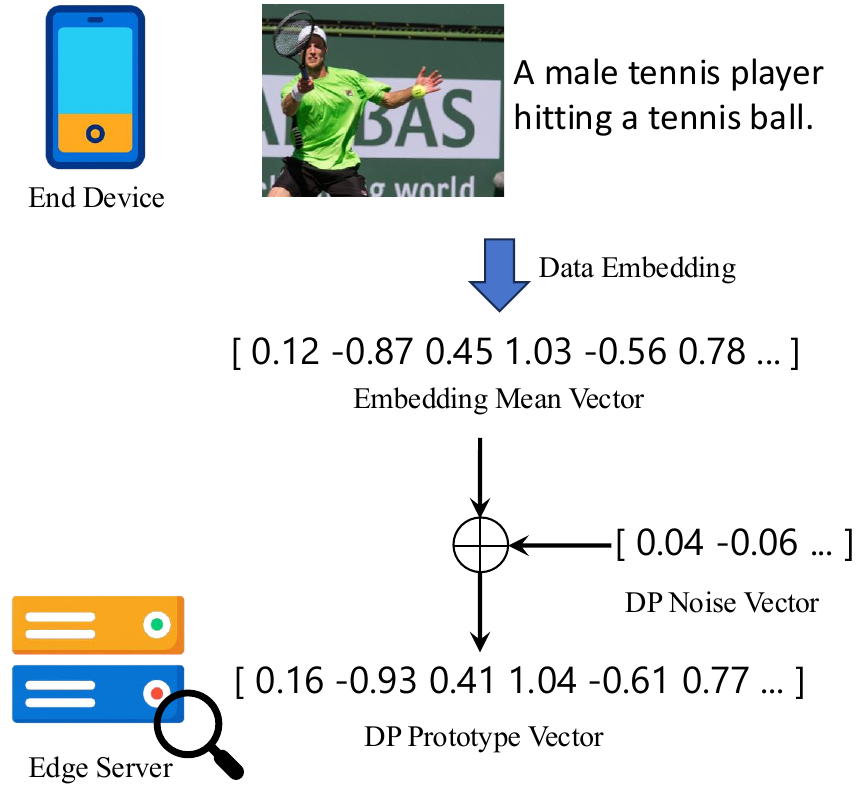}
		\caption{DP noise is added to high-dimensional private data before data valuation. Each semantic cluster forms an iDAO to govern data valuation and reward settlement.}
		\label{fig: dp_process}
	\end{figure}
    
    As shown in Fig.~\ref{fig: dp_process}, raw data (e.g., texts or images) are first encoded into numerical vectors in a semantic embedding space. As users continue to contribute new data, data embeddings form clusters in the semantic space. Data embeddings from the same cluster are averaged into a prototype. For privacy protection, DP noise is added to the prototype embedding vector before it is transmitted to the edge server.

    To incentivize high-quality contributions, we introduce a threshold $\Gamma$ to distinguish novel from normal data. Our intuition is that novelty is tied to timestamped data freshness: the first $\Gamma$ samples that populate a newly discovered cluster are treated as novel data, i.e., $\phi(\mathbf{d}_n) = 1$, while subsequent samples falling into the same cluster are considered normal contributions, i.e., $\phi(\mathbf{d}_n) = 0.5$. In contrast, noisy or low-quality data, which cannot be confidently assigned to any cluster, is regarded as semantically noisy and does not receive novelty credit, i.e., $\phi(\mathbf{d}_n) = 0$.

    Let $q$ denote the perturbation coefficient for the noise and $s$ denote a scale parameter for the Gaussian distribution. Let $\Delta_2$ denote $\ell_2$-sensitivity of the mean embedding. According to the Gaussian Mechanism~\cite{dwork2014algorithmic}, we require
    \begin{align}\label{equ:dp_gurantee}
        q^{2}s^{2} \;\geq\;
        \left(
        \frac{\Delta_2 \sqrt{2\ln(1.25/\delta)}}{\varepsilon}
        \right)^{2},
    \end{align}
    which guarantees the DP prototype obtained is $(\varepsilon,\delta)$-differentially private. Detailed derivations are included in the supplementary materials.

    \section{Results and Analysis}\label{sec: data_results}
    In this section, we evaluate our proposed schemes by computer simulations. In a decentralized edge environment, heterogeneous private data leads to diverse data contributions. We first use Jain's fairness index and Gini coefficient to test the fairness of AI quota allocation. Then, we evaluate our decentralized data valuation method in different blockchain environments. Finally, the proposed novelty detection with DP noise is evaluated for privacy protection.
    \subsection{Experimental Setup}
    We respectively set up three experiments for token allocation, data valuation, and privacy protection. The well-known COCO dataset~\cite{lin2014microsoft} is used as a multi-modal database in our simulations. The following benchmarks are used:
    \begin{itemize}
        \item Random allocation~\cite{zhang2025multimodal}: Distributes $V$ tokens randomly across $N$ agents according to a uniform multinomial distribution. Every token is independently assigned to one agent with equal probability, without bucket limitations.
        \item Round-Robin allocation~\cite{zhang2025multimodal}: Cycles through $N$ agents in a fixed order. Each agent is given one token at a time, capped by the bucket size and the available tokens.
        \item Max–Min allocation~\cite{sheng2024fairness}: Greedily gives the next token to the agent who currently holds the fewest tokens, capped by the bucket size and the available tokens.
        \item Image-DP\footnote{\url{https://anonymous.4open.science/r/DomainFL/}}~\cite{zhang2025enhancing}: Adds DP noise only to the image embeddings or prototypes. It protects visual features alone while leaving text embeddings unchanged. We follow~\cite{zhang2025enhancing} and set $q=0.2$, $s=0.05$.
        \item Reconstruction Attack\footnote{\href{https://github.com/stanislavfort/Direct_Ascent_Synthesis/}{https://github.com/stanislavfort/Direct\_Ascent\_Synthesis/}}~\cite{fort2025direct}: An adversary can reconstruct an image by optimizing a dummy input so its encoded embedding matches the shared prototype. We set $\epsilon$ = 1 and run 100 optimization steps with a learning rate of 0.2~\cite{fort2025direct}.
    \end{itemize}
    Unless otherwise indicated, we set $N=500$, $V=1000$, and $B=4$. We use the Dirichlet distribution with concentration parameter $\alpha = 0.5$ to simulate the heterogeneous distribution of data contributions. Our method (i.e., Proposed) is based on Equation (\ref{equ:user_token}). Each data point in the token allocation experiment is averaged across 1000 simulation rounds. Data embeddings are visualized and produced by using randomly selected images from the COCO validation set\footnote{\url{https://cocodataset.org/\#download}}. For decentralized data valuation, we set $\phi(\mathbf{d}_n)$ as described in Section~\ref{sec: novelty_valuation}. For multi-modal DP setting (i.e., Image-Ours and Text-Ours), we set $\epsilon=1$ and $\delta=10^{-5}$. For each newly discovered cluster, we use $\Gamma = 50$ as the threshold for the novel data quantity. The CLIP ViT/32 model~\cite{radford2021learning} is used across all experiments. Sensitivity analyses for $\alpha$ and $\Gamma$ are respectively provided in Fig.~\ref{fig: alpha} and Fig.~\ref{fig: de_value}. Further data visualizations are included in the supplementary materials.
    \subsection{Effectiveness of Token Allocation}
    A key challenge of token allocation is to balance the fairness of the reward rate $\mathbf{r}$ in the resource-limited network edge.
    \begin{figure}[t]
		\centering
		\subfigure[Jain Index]{
			\begin{minipage}{4.1cm}
				\centering
				\includegraphics[scale=0.35]{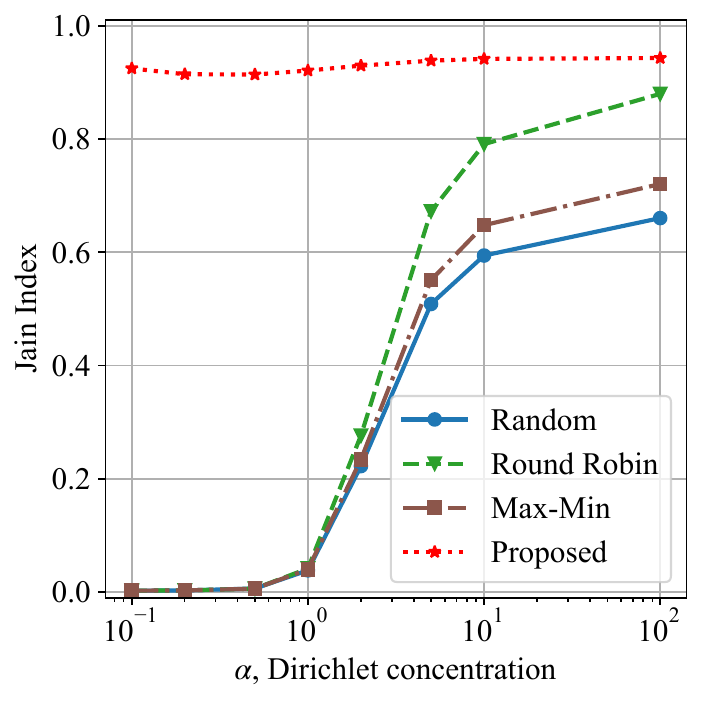}
			\end{minipage}
		}
		\subfigure[Gini Coefficent]{
			\begin{minipage}{4.1cm}
				\centering
				\includegraphics[scale=0.35]{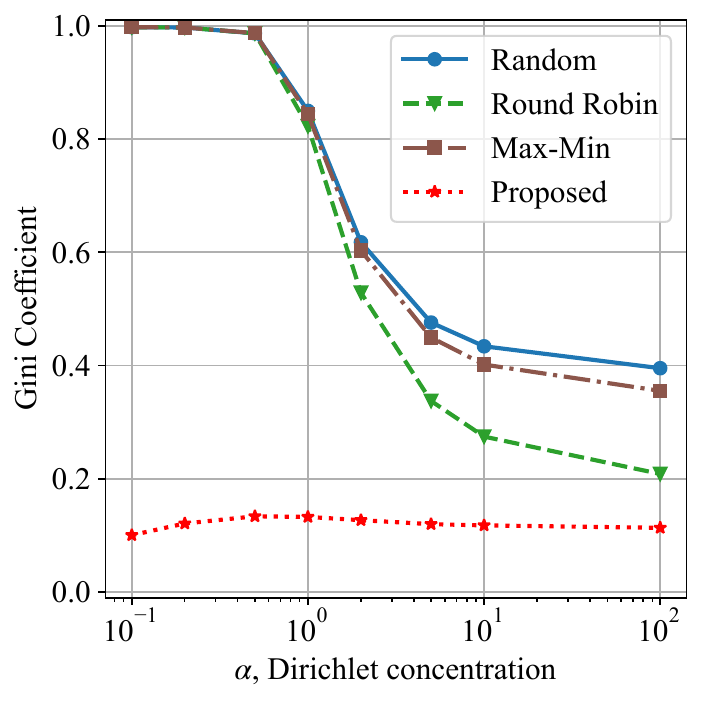}
			\end{minipage}
		}
		\caption{Fairness metrics under different $\alpha$. Our proposed token allocation improves fairness across diverse contribution distributions.}
		\label{fig: alpha}
	\end{figure}

    In Fig.~\ref{fig: alpha}, the distributional heterogeneity of $\mathbf{v}$ increases when $\alpha$ decreases. A higher Jain index and lower Gini coefficient lead to better fairness of AI quota allocation. A key observation is that our proposed token allocation scheme outperforms benchmarks across diverse data contribution scenarios. Our design achieves a ``contribute more to earn more" incentive mechanism.

    \begin{figure}[t]
		\centering
		\subfigure[Jain Index]{
			\begin{minipage}{4.1cm}
				\centering
				\includegraphics[scale=0.35]{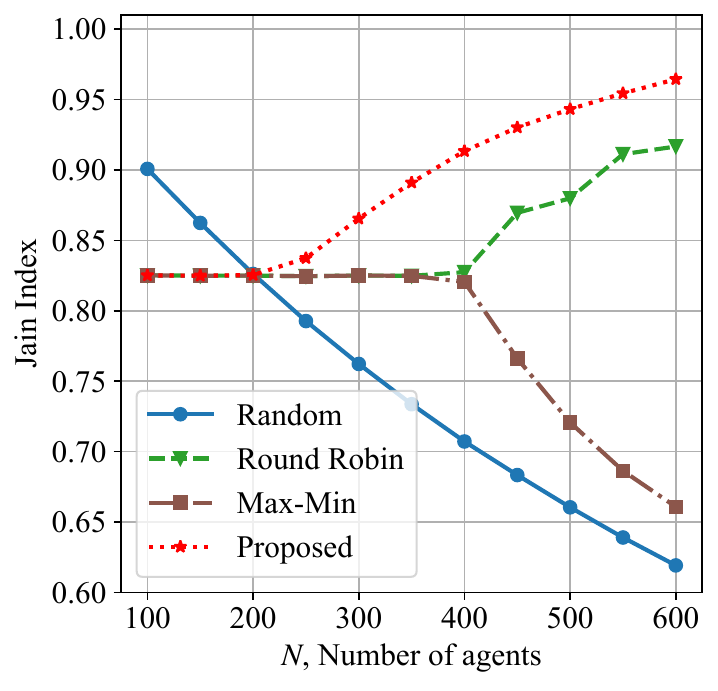}
			\end{minipage}
		}     
		\subfigure[Gini Coefficent]{
			\begin{minipage}{4.1cm}
				\centering
				\includegraphics[scale=0.35]{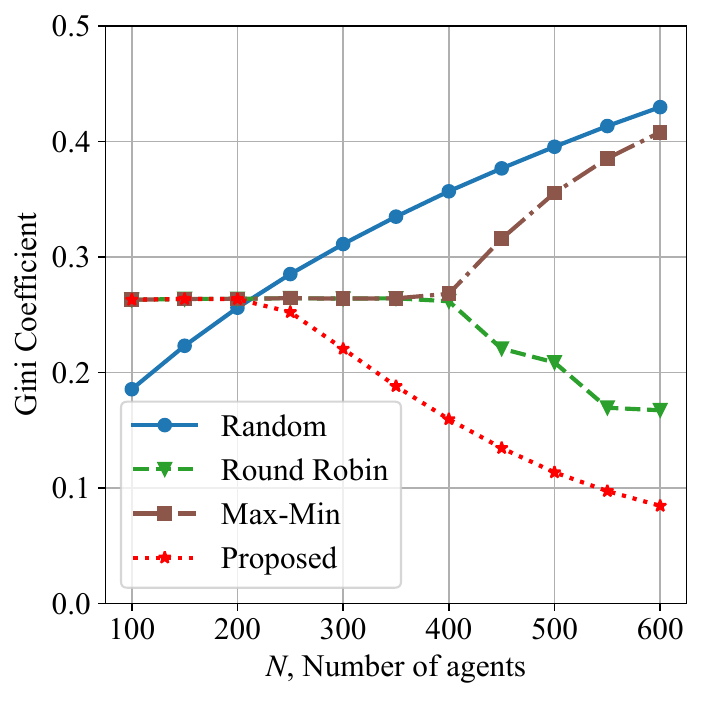}
			\end{minipage}
		}
		\caption{Fairness metrics with different $N$. Our method supports real-world deployment under constrained AI quota and compute budgets.}
		\label{fig: N}
	\end{figure}

    Fig.~\ref{fig: N} further illustrates the effectiveness of token allocation under varying numbers of agents $N$. With sufficient AI quota, all agents can be refilled to full capacity. However, in practical deployments where AI quota and computational resources are limited, our method significantly outperforms the benchmarks. The results indicate that our token allocation scheme is well-suited for large-scale distributed environments with constrained resources.
    \subsection{Decentralized Data Valuation}
    To show how novelty-based detection works, we use diverse $\Gamma$ to set the novelty threshold. The COCO dataset is used in the prototype discovery process with DP noise added.
    \begin{figure}[t]
		\centering
		\subfigure[Data Value]{
			\begin{minipage}{4.1cm}
				\centering
				\includegraphics[scale=0.35]{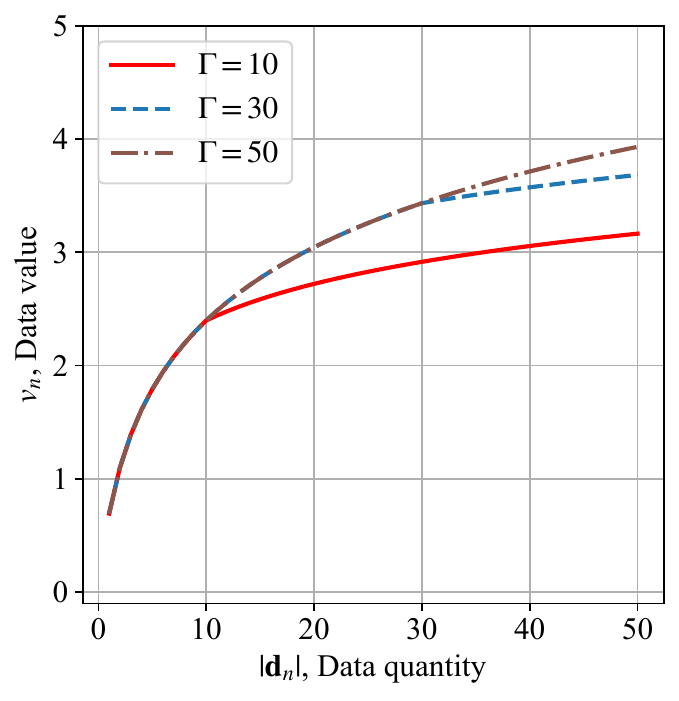}
			\end{minipage}
		}     
		\subfigure[Prototype Discovery]{
			\begin{minipage}{4.1cm}
				\centering
				\includegraphics[scale=0.35]{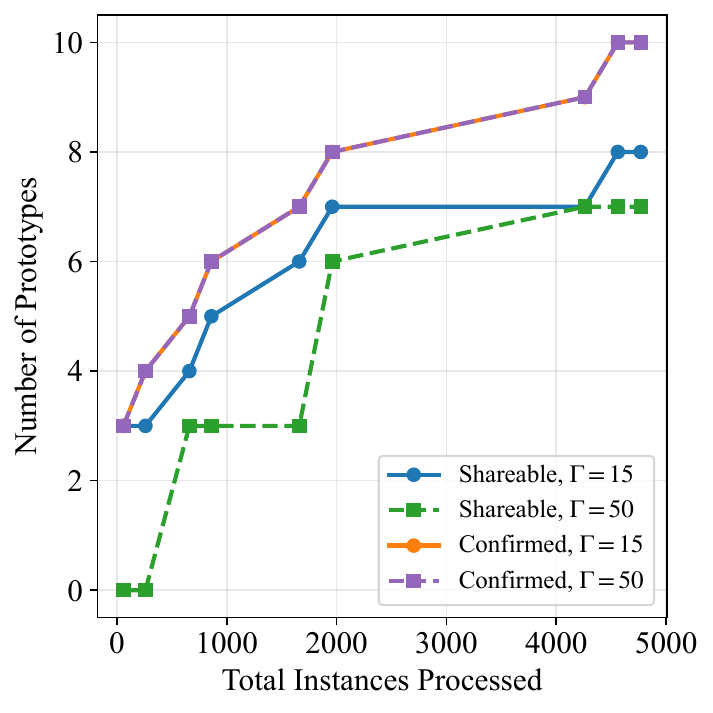}
			\end{minipage}
		}
		\caption{Data valuation in novel discovery. Our proposed scheme incentivizes early contributions while preserving privacy by applying DP before prototypes are shared.}
		\label{fig: de_value}
	\end{figure}

    As shown in Fig.~\ref{fig: de_value} (a), data valuation is governed by both quantity and novelty thresholds. The diminishing marginal utility incentivizes early contributions of novel data. In Fig.\ref{fig: de_value} (b), a larger $\Gamma$ admits more novel samples per cluster but delays cluster confirmation. Since DP noise is added only after a cluster is confirmed, increasing $\Gamma$ also postpones when DP can be applied. A trade-off between privacy and latency exists.
    \subsection{Multi-Modal Differential Privacy}
    To evaluate privacy protection, we conduct a reconstruction attack. The adversary uses the same CLIP encoder to optimize a dummy image so that its CLIP embedding maximizes similarity to the shared prototype.
    \begin{figure}[t]
    \centering
    \subfigure[Reconstructed Images]{%
        \begin{minipage}[t]{4.1cm}
            \centering
            \includegraphics[height=3.8cm,keepaspectratio]{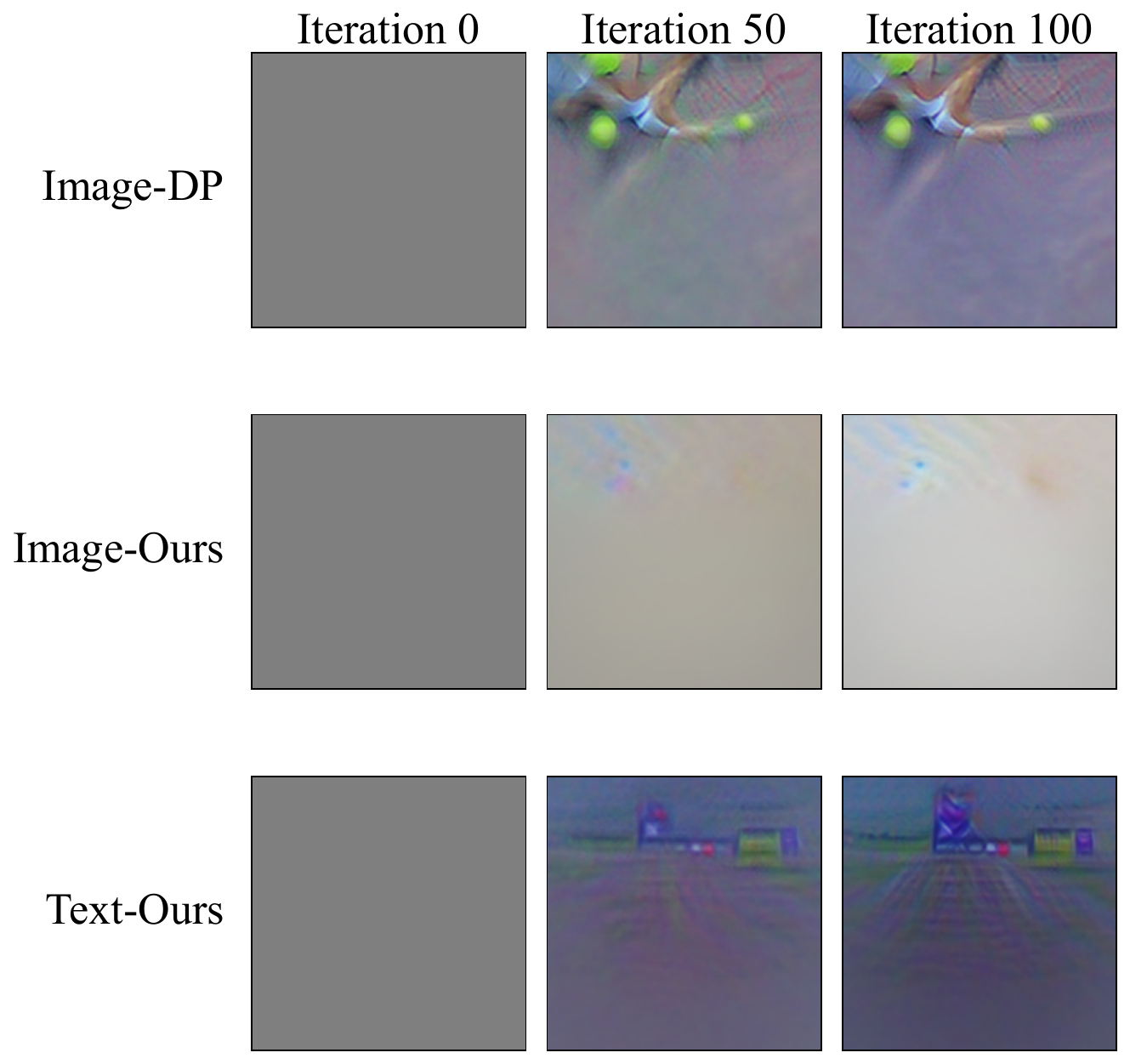}
        \end{minipage}%
    }\hspace{0.3cm}%
    \subfigure[Attack Score]{%
        \begin{minipage}[t]{4.1cm}
            \centering
            \includegraphics[height=3.8cm,keepaspectratio]{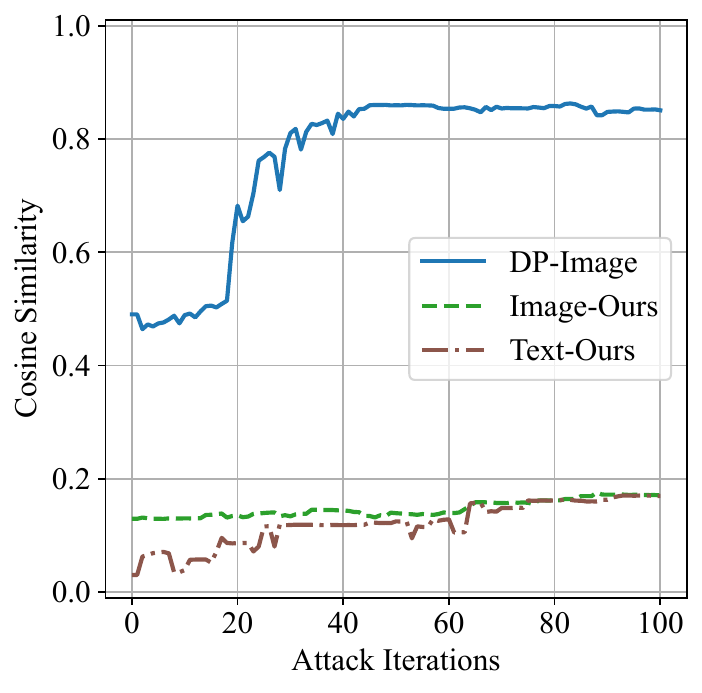}
        \end{minipage}%
    }
    \caption{Attacks on shared DP prototypes. Our method improves privacy level on multi-modal data.}
    \label{fig: attack}
    \end{figure}
    
    As illustrated in Fig.~\ref{fig: attack}, the Image-DP baseline leaks substantial semantic information. An example original sample is shown in Fig.~\ref{fig: dp_process}. In contrast, sensitive information, such as facial characteristics, is well-protected by our DP schemes. Our method achieves consistently lower attack scores than Image-DP. We observe that it is very challenging to reproduce the semantic information from our shared cross-modal prototypes. The results suggest that the DP guarantee in Equation~(\ref{equ:dp_gurantee}) strengthens privacy protection for multi-modal personal data.
    
    \section{Conclusions}\label{sec: conclusions}
    In this paper, we studied privacy-preserving data valuation for fair AI resource allocation in decentralized agentic systems. Our goal is to support data sovereignty by protecting multi-modal personal data with DP and enabling contribution-aware trading in decentralized data markets. We allocate limited AI quotas as tokens via a token-bucket allocator to achieve fair resource distribution under heterogeneous contributions. We further represent contributions using DP-protected prototypes, which form a semantic iDAO \emph{data catalog} (i.e., a marketplace directory) anchored in a shared embedding space. Despite the above advantages, the privacy–utility trade-off in releasing DP prototypes remains an open direction for future research.
    
    \bibliographystyle{IEEEtran}
    \bibliography{Reference}
    
    \section*{Supplementary Materials}
    
    \addcontentsline{toc}{section}{Supplementary Materials}
    
    \renewcommand{\thesection}{\arabic{section}}
    
    \renewcommand{\thesubsection}{S.\arabic{section}.\arabic{subsection}}
    
    \renewcommand{\thefigure}{S.\arabic{figure}}
    
    \renewcommand{\thetable}{S.\arabic{table}}
    
    \renewcommand{\theequation}{S.\arabic{equation}}
    
    \setcounter{section}{0}
    
    \setcounter{subsection}{0}
    
    \setcounter{figure}{0}
    
    \setcounter{table}{0}
    
    \setcounter{equation}{0}
    
    \section{Derivations of Differential Privacy Guarantee}
    For $I$ data instances in the same cluster, indexed by $i\in\mathcal{I} = \{1, 2, \dots, I\}$, let $\boldsymbol{\theta}_i$ denote the $i$-th embedding vector, clipped by $C$. That is, $\bigl\| \boldsymbol{\theta}_i \bigr\|_2 \leq C, \forall i\in\mathcal{I}$.
    
    Let $D$ and $D'$ be neighboring datasets differing in exactly one example, with $|D| = |D'| = I$. Let $\boldsymbol{\mu}(D) \in \mathbb{R}^d$, $\bar{\boldsymbol{\mu}}(D) \in \mathbb{R}^d$, and $\boldsymbol{\varsigma} \in \mathbb{R}^d$ respectively denote the embedding mean vector of $D$, the DP prototype of $D$, and the injected DP noise. Let $q$ denote the perturbation coefficient for the noise and $s$ denote a scale parameter for the Gaussian distribution. Let $\mathbf{I} \in \mathbb{R}^{d \times d}$ denote the identity matrix. A DP prototype is obtained by
    \begin{align}\label{equ:dp_process}
    	\bar{\boldsymbol{\mu}}(D)
    	= \boldsymbol{\mu}(D) + \boldsymbol{\varsigma}
    	= \frac{1}{I}\sum_{i=1}^{I}\boldsymbol{\theta}_i + \boldsymbol{\varsigma},
    \end{align}
    where $\boldsymbol{\varsigma}\sim\mathcal{N}\bigl(\mathbf{0}, q^{2}s^{2}\mathbf{I}\bigr)$~\cite{zhang2025enhancing}.
    
    The $\ell_2$-sensitivity of the mean embedding is
    \begin{align}
    	\Delta_2 
    	= \max_{D,D'} 
    	\bigl\| \boldsymbol{\mu}(D) - \boldsymbol{\mu}(D') \bigr\|_2 \leq \frac{2C}{I}.
    \end{align}
    
    According to the Gaussian Mechanism~\cite{dwork2014algorithmic}, we require
    \begin{align}
    	q^{2}s^{2} \;\geq\;
    	\left(
    	\frac{\Delta_2 \sqrt{2\ln(1.25/\delta)}}{\varepsilon}
    	\right)^{2},
    \end{align}
    which guarantees the DP prototype obtained is $(\varepsilon,\delta)$-differentially private. Equation (\ref{equ:dp_gurantee}) can be rewritten more explicitly as:
    \begin{equation}
    	q \cdot s \geq \frac{\Delta_2 \sqrt{2\ln(1.25/\delta)}}{\varepsilon}.
    \end{equation}
    
    Substituting the sensitivity bound $\Delta_2 \leq \frac{2C}{I}$:
    \begin{equation}\label{equ:final}
    	q \cdot s \geq \frac{2C\sqrt{2\ln(1.25/\delta)}}{I \cdot \varepsilon}.
    \end{equation}
    
    The product $q \cdot s$ must satisfy Equation (\ref{equ:final}) for $(\varepsilon,\delta)$-DP. We can fix either parameter and adjust the other to meet the privacy requirement.
    
    \section{Additional Reconstruction Attack Results}
    \begin{figure}[ht]
    	\centering
    	\includegraphics[width=\linewidth]{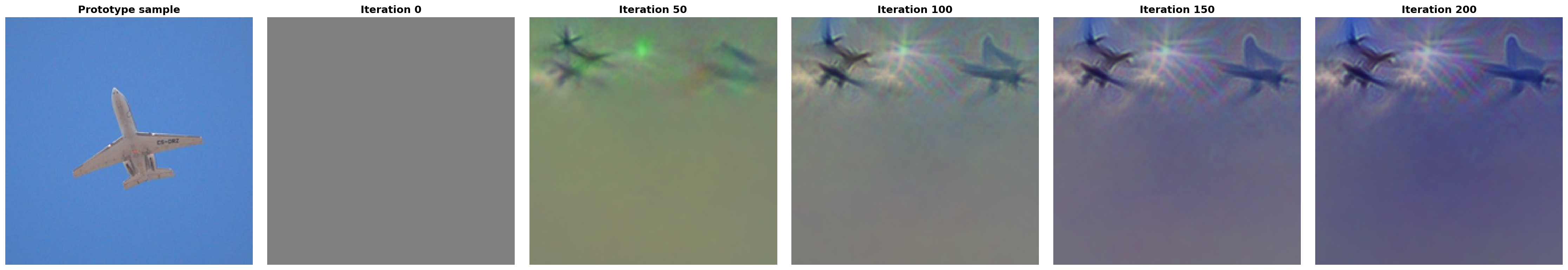}\par\vspace{0.6em}
    	\includegraphics[width=\linewidth]{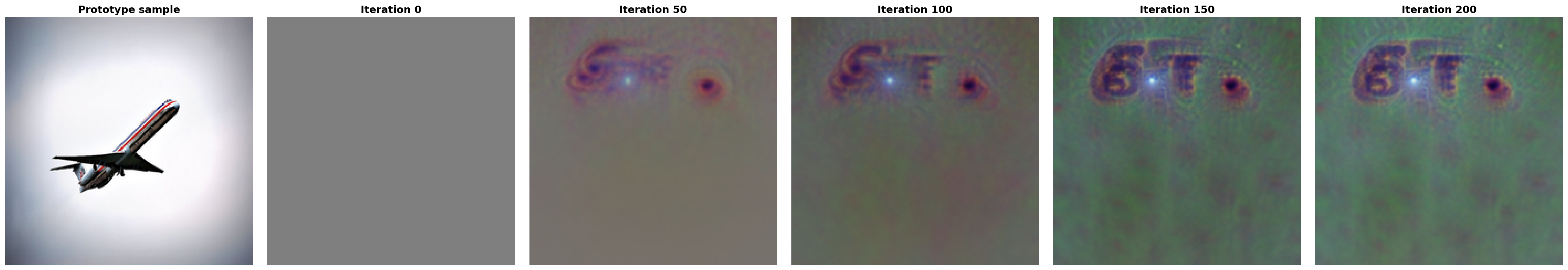}\par\vspace{0.6em}
    	\includegraphics[width=\linewidth]{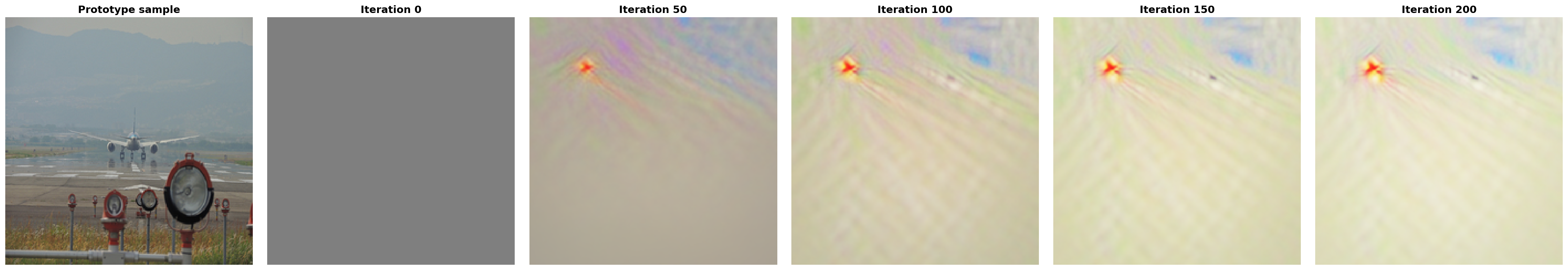}
    	\caption{Reconstruction attack comparison: (top) Image-DP benchmark~\cite{zhang2025enhancing} shows substantial semantic leakage; (middle) our Image-DP prototype is well-protected; (bottom) our Text-DP prototype is well-protected.}
    	\label{fig: attack_appendix_combined}
    \end{figure}
    
    We further provide additional experimental results in Fig.\ref{fig: attack_appendix_combined} to support the DP guarantee of our shared prototypes. Across 200 rounds of reconstruction attacks\cite{fort2025direct}, the benchmark method fails to prevent semantic leakage from the DP prototypes, whereas our proposed methods effectively protect the shared prototypes.
    
    \section{Dataset Visualization}

    \begin{figure*}[ht]
    	\centering
    	\includegraphics[width=0.75\linewidth]{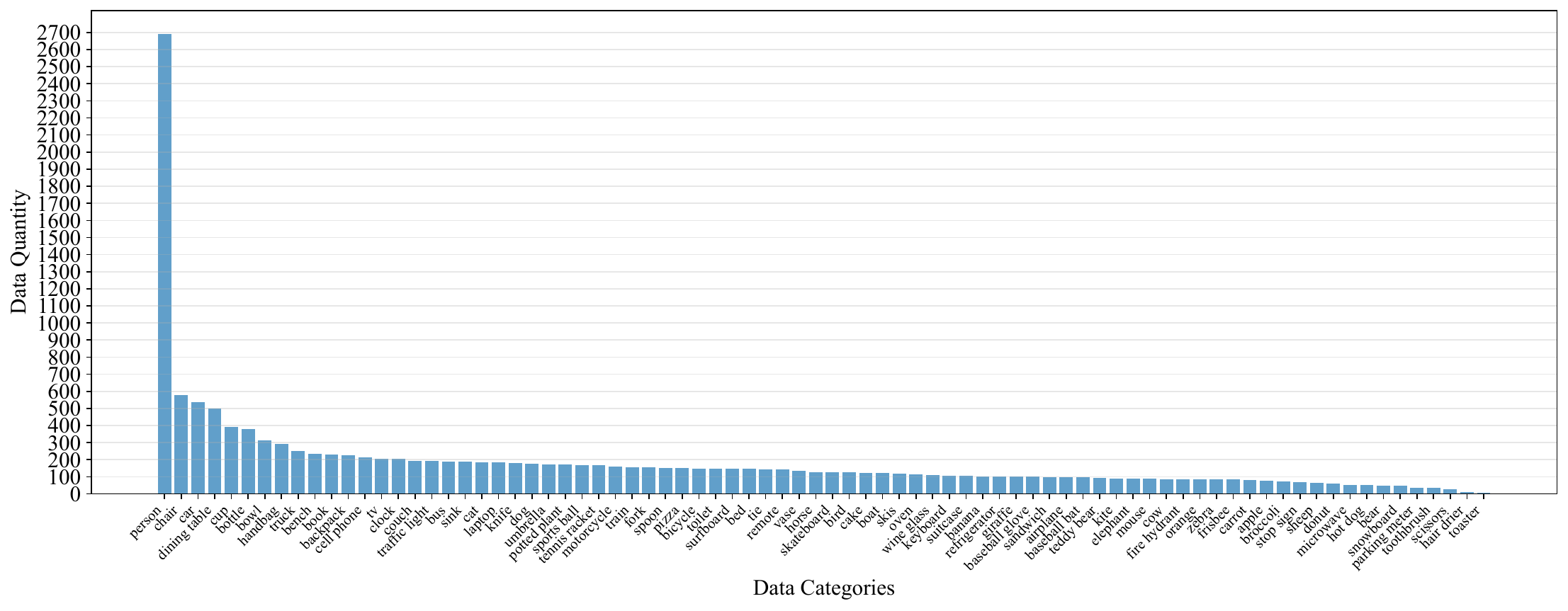}
    	\caption{Multi-modal datasets are highly skewed in quantity by nature: An example with the COCO validation dataset.}
    	\label{fig: coco_dataset}
    \end{figure*}
    
    \begin{figure}[ht]
    	\centering
    	\includegraphics[width=0.7\linewidth]{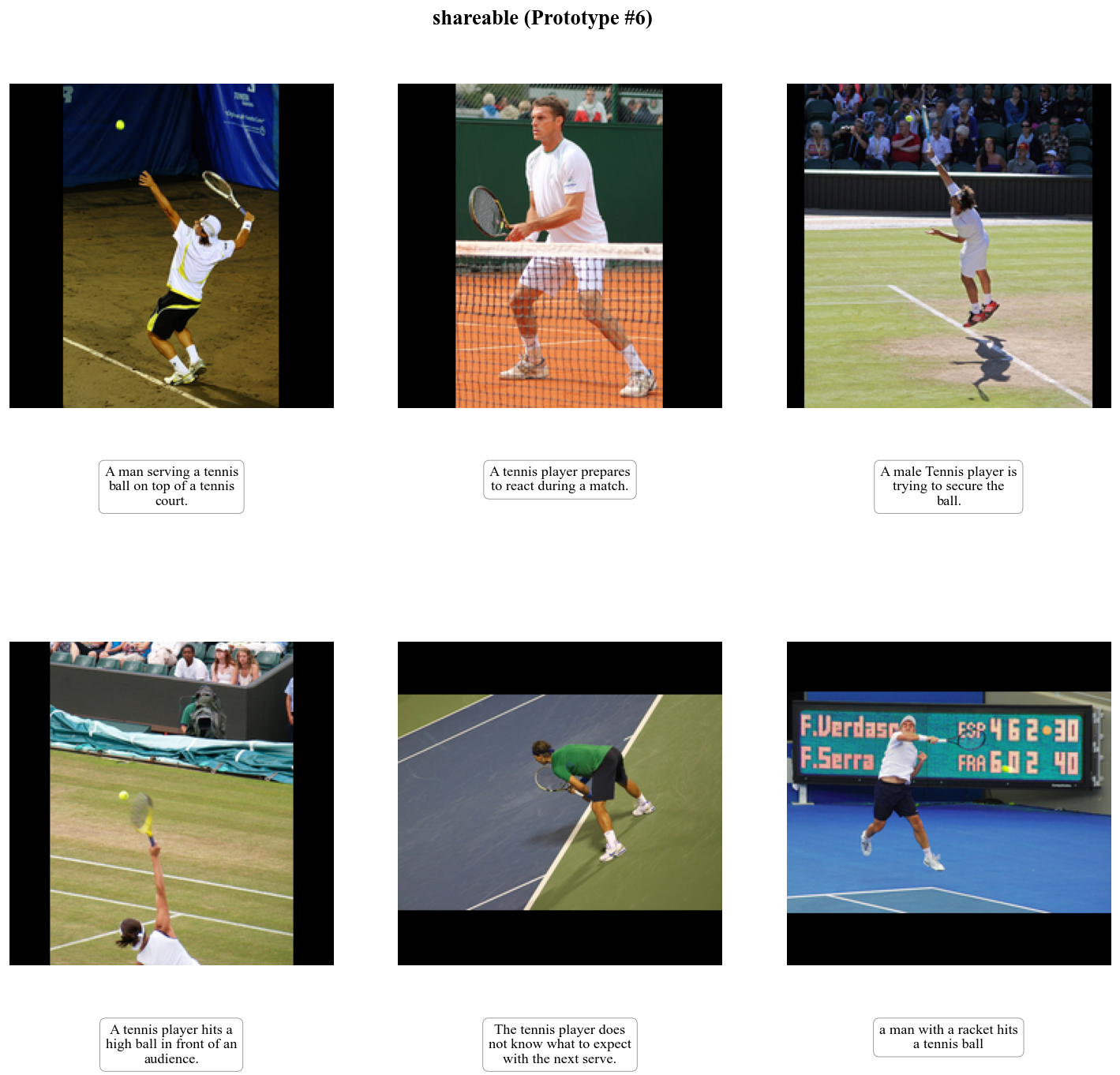}
    	\caption{Image samples from the shareable cluster: Detected cluster in the person class.}
    	\label{fig: shareable_6}
    \end{figure}
    
    \begin{figure}[ht]
    	\centering
    	\includegraphics[width=0.7\linewidth]{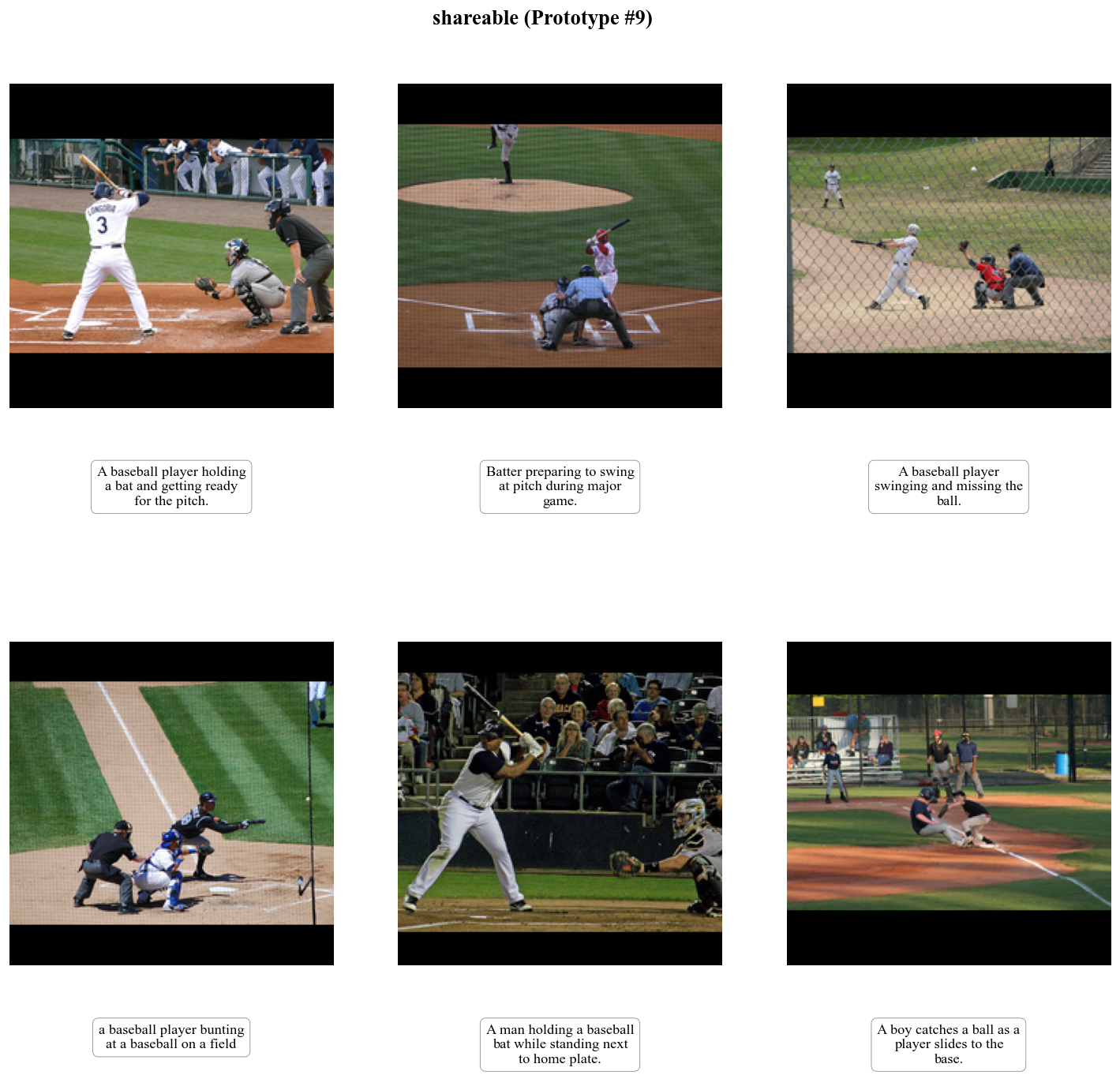}
    	\caption{Image samples from the shareable cluster: Detected cluster in the person class.}
    	\label{fig: shareable_9}
    \end{figure}
    
    We visualize the COCO dataset\footnote{\url{https://cocodataset.org/\#download}} in Fig.~\ref{fig: coco_dataset} to illustrate the severe data heterogeneity in decentralized settings. The COCO categories are dominated by the ``person" category, as human-related activities and objects are the most common subjects in data samples contributed from users.
    
    We further show sample images from the detected cluster in Fig.~\ref{fig: shareable_6} and Fig.~\ref{fig: shareable_9} through our novelty detection pipeline. While the COCO dataset does not explicitly label activities such as tennis or baseball, our novelty detection automatically clusters fine-grained semantic groups within the “person” category. Therefore, the proposed DP-protected detection process is effective for privacy-preserving knowledge discovery.
    
    \section{Limitations and Clarifications}
    
    \subsection{Bucket sizing and dynamic adjustment}
    The bucket size $B_n$ is determined by both the purchased service tier and the agent's data contribution level. Agents with higher tiers or greater contributions receive larger buckets and hence larger AI quota capacity. This design can also be extended to support dynamic adjustment of $B_n$ according to contribution and usage patterns. However, the principled design of such updates requires careful consideration of load balancing, fairness guarantees, and incentive compatibility. We therefore view dynamic bucket adaptation as an important direction for future work rather than a fully resolved component of the current system.
    
    \subsection{Privacy--utility trade-off}
    The encoder choice plays a central role in the privacy--utility trade-off. Larger CLIP encoders, such as ViT-L/14 compared with ViT-B/32, typically yield richer representations and higher-quality prototypes, which improve downstream utility. At the same time, these richer representations may retain more fine-grained information and thus increase susceptibility to reconstruction attacks. As a result, under the same $(\varepsilon,\delta)$ privacy budget, stronger utility may require stronger privacy protection. Our DP mechanism provides a principled framework for navigating this trade-off, but the optimal encoder choice remains application-dependent and is ultimately constrained by fundamental information-theoretic limits.
    
    \subsection{Optimization quality and computational overhead}
    The overall optimization outcome depends strongly on the quality of the initial token allocation stage. To improve robustness, we adopt the token-bucket mechanism rather than random token assignment, since random allocation is more likely to induce poor local minima and degrade the maximization of the social welfare objective. Although this design substantially reduces sub-optimality in practice, formal guarantees of global optimality are beyond the scope of the current work. 
    
    In addition, the proposed DP mechanism introduces client-side overhead because noise must be sampled and added to embeddings before transmission to edge servers for prototype aggregation. This procedure involves only lightweight arithmetic operations and scales linearly as $O(d)$ for $d$-dimensional embeddings. Nevertheless, a complete empirical evaluation of system overhead under diverse deployment conditions is deferred to future work.
    
    \subsection{Security assumptions and system resilience}
    Our system assumes normal Layer-2 execution for efficiency, while relying on Layer-1 for security enforcement, verification, and dispute resolution. In particular, Layer-1 smart contracts can verify Layer-2 behavior through fraud proofs or validity proofs, allowing malicious actions to be detected and penalized. This provides accountability even when Layer-2 servers behave incorrectly. However, if a significant fraction of Layer-2 nodes are compromised, the system may still suffer temporary service disruption or degraded availability before Layer-1 resolution is finalized. Therefore, while adversarial behavior is ultimately challengeable and punishable, short-term resilience and service continuity remain important limitations of the current design.
\end{document}